\documentclass[10pt,twocolumn,letterpaper]{article}

\usepackage{cvpr}              

\newcommand{\thename}[0]{Senna-2}

\def\eg{\emph{e.g.}}
\def\ie{\emph{i.e.}}

\newcommand{\tablestyle}[2]{\setlength{\tabcolsep}{#1}\renewcommand{\arraystretch}{#2}\centering\footnotesize}
\usepackage{pifont}
\usepackage{multirow}
\usepackage[table]{xcolor}
\usepackage{makecell}
\usepackage{bbm}
\usepackage{marvosym}

\definecolor{cvprblue}{rgb}{0.21,0.49,0.74}
\usepackage[pagebackref,breaklinks,colorlinks,allcolors=cvprblue]{hyperref}

\def\paperID{****} 
\def\confName{CVPR}
\def\confYear{2026}

\title{\thename{}: Aligning VLM and End-to-End Driving Policy for  \\  Consistent Decision Making and Planning}

\author{
\textbf{Yuehao Song}$^{1}$ \quad
\textbf{Shaoyu Chen}$^{2,\dagger}$ \quad
\textbf{Hao Gao}$^{1}$ \quad  
\textbf{Yifan Zhu}$^{2}$ \quad 
\textbf{Weixiang Yue}$^{2}$ \quad 
\textbf{Jialv Zou}$^{1}$ \quad  \\
\textbf{Bo Jiang}$^{1}$ \quad  
\textbf{Zihao Lu}$^{2}$ \quad 
\textbf{Yu Wang}$^{2}$ \quad 
\textbf{Qian Zhang}$^{2}$ \quad 
\textbf{Xinggang Wang}$^{1,\textrm{\Letter}}$ \\
\textsuperscript{1}\,Huazhong University of Science \& Technology \\
\textsuperscript{2}\,Horizon Robotics \\
\normalsize{
\url{https://ambitious-idiot.github.io/senna2-project}
}\\
\normalsize{
\url{https://github.com/hustvl/Senna}
}
}

\begin{document}
\maketitle

\renewcommand{\thefootnote}{\relax}
\footnotetext[2]{$^\dagger$ Project leader. $^\textrm{\Letter}$ Corresponding author.}
\renewcommand{\thefootnote}{\arabic{footnote}}

\begin{abstract}

Vision-language models (VLMs) enhance the planning capability of end-to-end (E2E) driving policy by leveraging high-level semantic reasoning.
However, existing approaches often overlook the dual-system consistency between VLM's high-level decision and E2E's low-level planning.
As a result, the generated trajectories may misalign with the intended driving decisions, leading to weakened top-down guidance and decision-following ability of the system.
To address this issue, we propose \thename{}, an advanced VLM-E2E driving policy that explicitly aligns the two systems for consistent decision-making and planning.
Our method follows a consistency-oriented three-stage training paradigm.
In the first stage, we conduct driving pre-training to achieve preliminary decision-making and planning, with a decision adapter transmitting VLM decisions to E2E policy in the form of implicit embeddings.
In the second stage, we align the VLM and the E2E policy in an open-loop setting.
In the third stage, we perform closed-loop alignment via bottom-up Hierarchical Reinforcement Learning in 3DGS environments to reinforce the safety and efficiency.
Extensive experiments demonstrate that \thename{} achieves superior dual-system consistency (19.3\% F1 score improvement) and significantly enhances driving safety in both open-loop (5.7\% FDE reduction) and closed-loop settings (30.6\% AF-CR reduction).\vspace{-0.2em}

\end{abstract}
\section{Introduction}
\label{sec:intro}

Reliable autonomous driving systems are expected to integrate high-level decision making as guidance for low-level trajectory planning.
Ensuring such decision–planning consistency enables the generation of trajectories that faithfully reflect driving intentions.
Recent end-to-end (E2E) driving frameworks~\cite{uniad,vad,vadv2,genad,diffusiondrive,goalflow,transfuser} have demonstrated strong capability in mapping sensory inputs to driving plans through unified perception–prediction–planning pipelines.
However, they still fall short in high-level reasoning and decision-making, which are crucial for ensuring safety and efficiency in complex driving scenarios.
Meanwhile, vision–language models (VLMs)~\cite{internvl,internvl3_5,qwenvl,qwen2_5vl,infinitevl} exhibit powerful abilities in scenario understanding and causal reasoning.
Recent studies~\cite{senna,opendrivevla} have integrated the VLM into the E2E policy to effectively remedy this cognitive defect.

\begin{figure}[t]
    \centering
    \includegraphics[width=0.9\columnwidth]{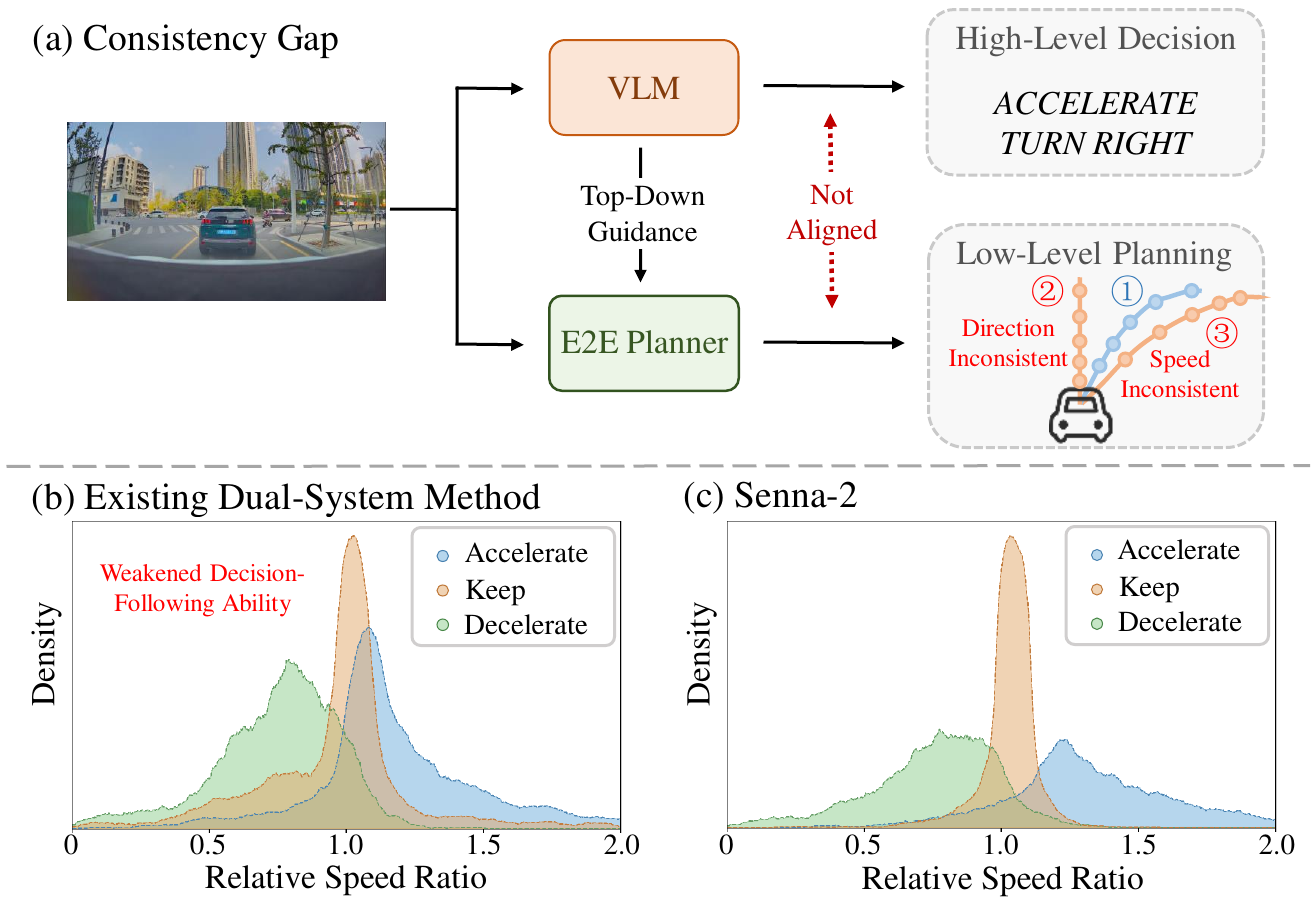}
    \caption{\textbf{Consistency gap between VLM and the E2E planner.}
    (a) The E2E planner may misalign with the VLM decision (\ding{172}), \eg, \ding{173}wrong direction or \ding{174}mismatched speed change.
    (b) Existing method~\cite{senna} shows scattered speed distributions, inconsistent with speed decisions.
    (c) With consistency-oriented training, \thename{} produces more distinct and decision-aligned speed distributions, reflecting improved dual-system consistency and decision-following ability.
    Relative speed ratio: the ratio between the planned speed at the 3rd second and the initial speed, reflecting the tendency of speed change of the planning trajectory.}\vspace{-0.2em}
    \label{fig:teaser}
\end{figure}

Despite recent advances in VLM-E2E driving policies~\cite{realad,diffvla,orion}, a significant consistency gap remains between high-level decision making and low-level trajectory planning.
As illustrated in~\cref{fig:teaser}(a), existing methods typically use VLM decisions to guide the planning of the E2E policy.
However, without explicit alignment between them, the resulting trajectories often deviate from driving intentions and lead to incorrect driving directions or mismatched speed changes.
Such inconsistency weakens the top-down guidance ability of the VLM and interpretability, ultimately causing suboptimal or even unsafe driving behavior.

We identify the root cause as the absence of \textbf{explicit alignment between high-level decision making and low-level planning}.  
To address the issue,
we develop a three-stage consistency-oriented training framework that aligns the VLM's high-level decision making with the E2E planner's low-level planning, ensuring consistent decision making and planning across the two systems:
\begin{enumerate}
    \item \textit{Driving pre-training}: The VLM and E2E policy are trained on large-scale driving data to achieve preliminary ability of decision making and planning, while the decision adapter is optimized to bridge the two systems.
    \item \textit{Open-loop alignment}:
    We identify inconsistencies between the VLM and the E2E policy by analyzing the kinematic discrepancies between the planned trajectories and the corresponding VLM decisions, and selectively refine these cases to enhance dual-system consistency.
    \item \textit{Closed-loop alignment with Hierarchical Reinforcement Learning (HRL)}: We further optimize the system in closed-loop 3DGS environments~\cite{rad} via online HRL. To balance safety and efficiency, we design composite rewards that jointly consider both aspects.
    Based on these rewards, the policy is optimized hierarchically: we first refine the E2E planner through longitudinal scaling penalties, and then update the VLM accordingly.
\end{enumerate}

Built upon this training pipeline, we introduce \textbf{\thename{}}, a unified VLM–E2E driving system that ensures consistent decision-making and planning.
\thename{} integrates the semantic reasoning power of the VLM with the fine-grained planning capability of the E2E policy, leading to safer and more coherent driving behaviors.
This design offers two key advantages:
(1) The VLM acts as an interpretable human–machine interface that both regulates driving behavior and explains high-level decisions.
(2) The combination of the VLM’s strong generalization and the E2E policy’s adaptability enhances robustness against long-tail scenarios, achieving more consistent and controllable planning compared with existing methods~\cite{senna} (\cref{fig:teaser}(b, c)).

We evaluate \thename{} across multiple dimensions, including decision-planning consistency, open-loop planning performance, and closed-loop driving stability.
Experimental results demonstrate that \thename{} achieves superior consistency between decision making and planning (19.3\% F1 score improvement compared to Senna~\cite{senna}),  
while maintaining competitive accuracy and robustness in both open-loop benchmarks (final displacement error reduction by 5.7\%) and closed-loop benchmarks (at-fault collision rate reduction by 30.6\%).

The contributions are summarized as follows:
\begin{itemize}
    \item We present \thename{}, a unified VLM-E2E driving policy that achieves consistent decision making and planning.
    \item We propose a consistency-oriented three-stage training paradigm that progressively aligns high-level decisions and low-level planning via large-scale pre-training, open-loop alignment, and closed-loop alignment with HRL.  
    \item Experiments on large-scale driving benchmarks validate that \thename{} significantly improves dual-system consistency, planning accuracy, and closed-loop robustness.
\end{itemize}
\section{Related Work}
\label{sec:related_work}

\subsection{End-to-End Autonomous Driving}

End-to-end autonomous driving policies evolve from unified perception–planning frameworks toward probabilistic, generative, and closed-loop consistent policies.
UniAD~\cite{uniad} and VAD~\cite{vad} pioneer unified architectures that integrate perception, prediction, and planning to realize trajectory reasoning directly from sensor inputs.
SparseDrive~\cite{sparsedrive} further explores sparse representations to improve efficiency without sacrificing global context.
VADv2~\cite{vadv2} and the Hydra-MDP series~\cite{hydramdp,hydramdp++} extend the paradigm with multi-modal planning to enhance behavioral diversity and robustness.
Generative methods~\cite{genad,goalflow,transdiffuser,diffusiondrive,diffusiondrivev2} model the trajectory distribution as a generation task and yield more expressive driving policies beyond deterministic regression.
Meanwhile, RAD~\cite{rad} utilizes closed-loop optimization to narrow the gap between open-loop training and real-world deployment.
Despite these advances, most end-to-end frameworks lack high-level reasoning, motivating the integration of vision–language models for more controllable and explainable decision-making.

\begin{figure*}[!t]
    \centering
    \includegraphics[width=0.93\textwidth]{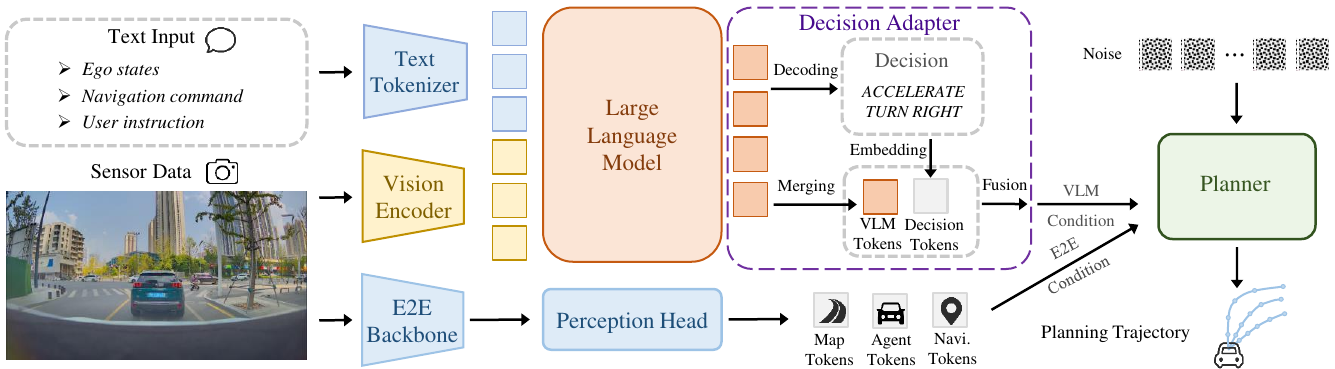}
    \caption{\textbf{Overall model architecture of \thename{}.}
    Text and visual inputs are processed by the VLM to produce high-level driving decisions, which are converted by the Decision Adapter into VLM condition embeddings. The E2E planner then fuses the VLM condition with its own E2E features to generate a trajectory consistent with the high-level decisions.}\vspace{-0.8em}
    \label{fig:framework}
\end{figure*}

\subsection{VLM for Autonomous Driving}

Recent progress in vision–language models (VLMs) has introduced a new paradigm for autonomous driving.
Early work~\cite{vlaad,hintad,drivegpt4,dynrslvlm} formulates driving as a language-centric problem and employs VLMs for scene understanding through question answering.
Another line of research~\cite{emma,openemma,opendrivevla,recogdrive,lmdrive,drivevlm,omnidrive} endows VLMs with trajectory generation capability by directly integrating planning into the VLM architecture.
However, as the semantic and action spaces are inherently heterogeneous\cite{pi0.5}, such tight coupling often leads to unstable optimization.
To address this, Senna~\cite{senna} decouples high-level decision making from low-level planning and leverages VLM decisions to guide end-to-end policies.
Subsequent works further enhance this paradigm by introducing explicit reasoning processes~\cite{alphadrive,autovla} or hierarchical instructions~\cite{realad}.
More recent approaches incorporate generative planners~\cite{diffvla,orion} to synthesize diverse, context-aware trajectories guided by high-level decisions.
Nevertheless, current VLM-based driving systems still lack explicit consistency constraints between the two systems, which leads to potential misalignment between decision making and planning.

\subsection{Cross-System Consistency in Driving}

While VLM-based frameworks introduce high-level decisions to driving systems, it remains challenging to maintain consistency between decision making and planning.
SimLingo~\cite{simlingo} proposes an action dreaming task to evaluate the alignment between instructions and planning.
VLM-AD~\cite{vlmad} and ALN-P3~\cite{alnp3} propose feature alignment between the perception, prediction, and planning features of the VLM and the E2E policy.
RDADriver\cite{rdadriver} proposes a reasoning-decision alignment constraint between the paired CoTs and planning results.
However, these approaches mainly focus on data-level or representation-level alignment and lack explicit consistency constraints between decision making and planning.

\section{Method}
\label{sec:method}

\subsection{Dual-System Architecture}
\label{sec:model_architecture}

Our model architecture consists of three components: the VLM module, the decision adapter, and the end-to-end driving policy, as illustrated in~\cref{fig:framework}.
The VLM module generates the high-level decisions, the decision adapter translates them into conditioning features, and the end-to-end policy produces final driving planning guided by these features.

\paragraph{VLM.}
We adopt Qwen2.5-VL-3B~\cite{qwen2_5vl} as the base model for high-level driving decision-making to achieve a good trade-off between performance and efficiency.
The input consists of a single front-view frame and the text input, including the system prompt, a navigation command, and the ego speed.
After tokenization, the visual and textual tokens are jointly fed into the language model to predict the driving decision.
The driving decision is composed of speed control and direction control.
Speed control includes \textit{acceleration}, \textit{deceleration}, \textit{keep speed}, and \textit{stop}. Direction control covers \textit{go straight}, \textit{turn left}, \textit{turn right}, \textit{change lane left}, and \textit{change lane right}.
These structured meta actions serve as interpretable decisions that guide the trajectory planning.

\paragraph{Decision Adapter.}
We design a decision adapter to transform high-level decisions produced by the VLM into representations compatible with the end-to-end driving model.
Specifically, the adapter outputs two complementary types of tokens: \textbf{VLM tokens} and \textbf{decision tokens}.
We extract the final hidden states from the VLM and project them through an MLP to obtain the VLM tokens $T_\mathrm{vlm}$, which preserves the reasoning context from the VLM.
Meanwhile, to enhance the model’s awareness of VLM decisions, we introduce learnable category embeddings for speed and direction control.
According to the decoded meta-actions, the velocity embedding $T_\mathrm{vel}$ and the direction embedding $T_\mathrm{dir}$ are selected as decision tokens, respectively.
The VLM tokens and the two decision tokens are then fused to form the VLM condition feature:
\begin{equation}
F_\mathrm{vlm}=\mathrm{MLP}\big(\mathrm{Concat}(T_\mathrm{vlm},T_\mathrm{vel},T_\mathrm{dir})\big).
\end{equation}
This fusion design combines the expressiveness of VLM tokens in capturing global semantics with the structured interpretability of VLM decisions.

\begin{figure*}[!t]
    \centering
    \includegraphics[width=0.93\textwidth]{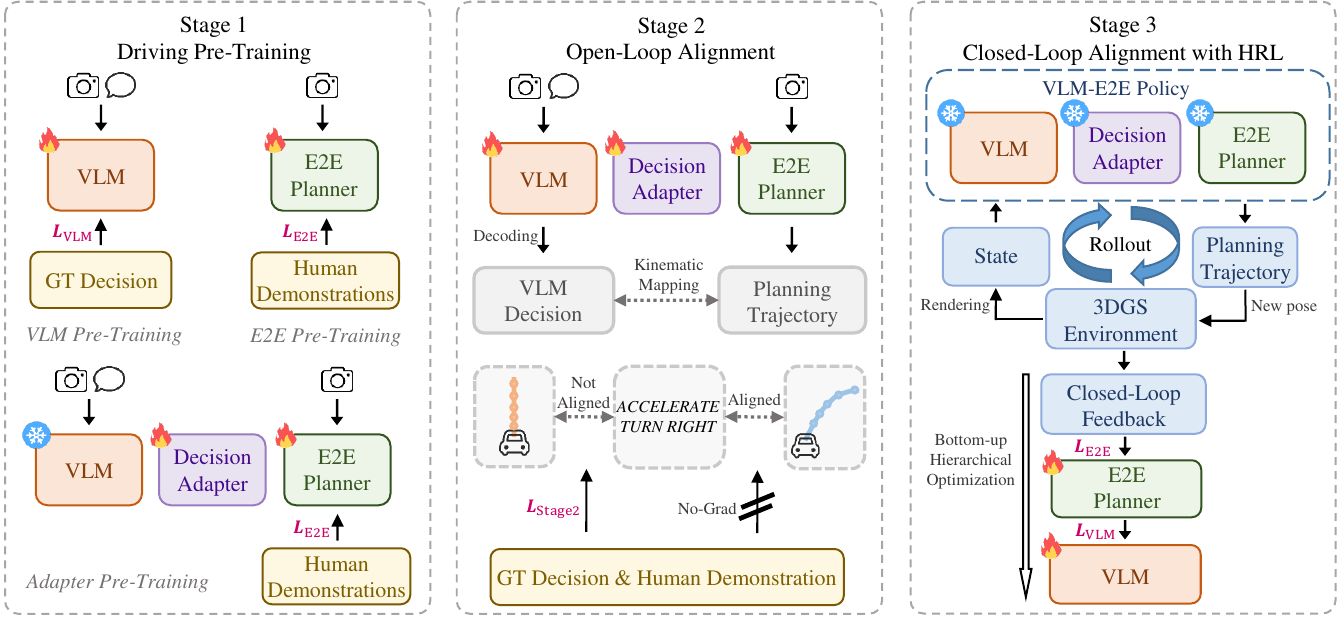}
    \caption{\textbf{Consistency-oriented training recipe.} We perform three training stages, including driving pre-training, open-loop alignment, and closed-loop alignment with Hierarchical Reinforced Learning (HRL).}\vspace{-0.8em}
    \label{fig:training}
\end{figure*}

\paragraph{End-to-End Driving Policy.}

Our end-to-end driving policy (E2E policy) consists of an E2E backbone, multiple perception heads, and a diffusion-based planner.
The E2E backbone extracts spatio-temporal features from multi-view image sequences.
The perception heads then predict features of static map elements, dynamic agents, and navigation cues based on the spatio-temporal features and navigation commands.
These features are concatenated to form the overall E2E condition.
The planner employs a DiT architecture~\cite{dit} and incorporates the E2E condition through cross attention~\cite{transformer}.
To introduce high-level decision guidance, we inject the VLM condition into the planning network using AdaLN~\cite{dit}.
This mechanism enables the VLM condition to globally modulate the planning process, guiding the generated trajectories to align with high-level decisions.

\subsection{Training Recipe}
\label{sec:training}

We adopt a three-stage training paradigm: driving pre-training, open-loop alignment, and closed-loop alignment with HRL, as shown in \cref{fig:training}.

\subsubsection{Driving Pre-Training}
\label{sec:stage1}

We first perform driving pre-training to equip the VLM with basic driving knowledge and align it with the E2E policy.

\paragraph{VLM Pre-Training.}
\label{sec:stage1.1}

We define a kinematic mapping function $f_K$ that converts ground-truth trajectories into meta-actions $d$ composed of speed and direction components, following the design in Senna~\cite{senna}.
This meta-action text serves as the answer in a question–answering (QA) formulation.
Meanwhile, the front-view image, navigation text, and ego state are used to construct the question.
We train the VLM through QA-based supervision:
\begin{equation}
    \mathcal{L}_{\mathrm{VLM}}=-\sum \log P(d_t\mid d_{< t},Q),
\end{equation}
where $Q$ denotes the question input and $d_t$ represents the $t$-th meta-action token.

\paragraph{E2E Pre-Training.}
\label{sec:stage1.2}

We model the trajectory generation as a diffusion-based planning problem~\cite{diffusiondrive}.
We set the prediction target as the residual trajectory~\cite{resad}, \ie, the difference between the expert trajectory and the reference trajectory extrapolated from the initial speed.
The planner takes the perception tokens as a condition and predicts this residual trajectory through a diffusion denoising process.
During training, we adopt the standard diffusion loss between the predicted and ground-truth noise:
\begin{equation}
    \mathcal{L}_{\mathrm{E2E}}=\mathbb{E}_{\mathbf{r}_0,\epsilon,t}[\left\lVert\epsilon-\epsilon_{\theta}(\mathbf{r}_t,t,c)\right\rVert^2],
\end{equation}
where $\mathbf{r}_0$ is the ground-truth residual trajectory, $\mathbf{r}_t$ is the noisy sample at time step $t$, $c$ represents the E2E condition tokens, $\epsilon$ is the ground-truth noise, and $\epsilon_\theta$ is the predicted noise produced by the E2E model parameterized by $\theta$.

\paragraph{Adapter Pre-Training.}
\label{sec:stage1.3}

After standalone training, we introduce the decision adapter.
We freeze the VLM parameters and optimize both the decision adapter and the E2E policy solely using the end-to-end planning loss $\mathcal{L}_{\mathrm{E2E}}$.
This ensures that the adapter can bridge the modules without affecting the decision-making capabilities of the VLM.

\subsubsection{Open-Loop Alignment}
\label{sec:stage2}
In the second stage, we adopt an open-loop alignment training scheme.
The core idea is to use the consistency between the VLM decision and the E2E planning as an explicit optimization signal. When the two systems agree, the model reinforces the corresponding behaviors; otherwise, it updates the policy under external supervision to correct the gap.

Specifically, we employ a kinematic mapping function $f_K$ to project the predicted trajectory $\tau$ into its corresponding decision category, which is then compared with the VLM decision $d$ to assess their behavioral consistency.
To formalize this, we define a simple yet effective binary consistency indicator:
\begin{equation}
\mathcal{C}(\tau,d) = \begin{cases}1&f_K(\tau)=d,\\0 & otherwise.\end{cases}
\end{equation}
The two are considered consistent when they belong to the same category, and inconsistent otherwise.

For inconsistent samples, the mismatch between high-level decisions and low-level planning leads to unreliable behaviors.
To maintain the rationality of planning while reducing the consistency gap, we apply explicit supervision using expert trajectories and corresponding decision labels.
In contrast, for consistent samples, the model regards its prediction as an internally coherent behavior.
Inspired by negative sample reinforcement~\cite{nsr}, we skip external supervision for these samples and treat the prediction itself as an implicit expert signal that provides self-reinforcing feedback.

Based on this mechanism, the loss function is defined as:
\begin{equation}
\mathcal{L}_\mathrm{stage2}=(1-\mathcal{C}(\tau,d))(\mathcal{L}_\mathrm{E2E}+\gamma \mathcal{L}_\mathrm{VLM}),
\label{eq:stage2}
\end{equation}
where $\gamma$ is a balancing weight.
This adaptive training strategy allows the model to achieve higher consistency through dynamic supervision.

\subsubsection{Closed-Loop Alignment with HRL}
\label{sec:stage3}
While open-loop alignment ensures policy consistency in regular scenarios, its reliance on offline supervision limits performance in out-of-distribution situations~\cite{raft}.
To address this, we introduce a closed-loop alignment via online Hierarchical Reinforcement Learning (HRL), which better aligns decisions and planning, thereby enhancing both safety and efficiency in real-world scenarios.

To enable closed-loop training, we first collect a large set of high-risk, dense-traffic clips from driving demonstrations.
Each clip is converted into an independent digital driving environment using 3D Gaussian Splatting (3DGS)~\cite{rad}.
During training, the VLM–E2E policy is deployed in a subset of environments to control the ego vehicle and generate rollouts.
These rollouts are then used to compute rewards and optimize the policy.
Optimization follows a bottom-up hierarchical scheme: the low-level planner is first optimized using safety and efficiency rewards, and the resulting improvements are propagated to update the high-level decision, ensuring consistent alignment between the two systems.

\paragraph{Low-Level Planner Reward Design.}
For the low-level planner, we design two complementary rewards based on the positional and motion states of the ego and surrounding vehicles.
The safety reward is computed from the time-to-collision (TTC), where trajectories are penalized in the longitudinal direction if a collision risk is detected ($\mathrm{TTC} < 3s$) to discourage unsafe behaviors.
The efficiency reward is applied when the ego speed is substantially below both the navigation speed limit and the reference demonstration speed.
These rollouts are treated as low-efficiency samples and are encouraged using longitudinal extension to promote smoother and faster driving. Together, these rewards guide the low-level planner to balance caution with efficiency during closed-loop training.
The above mechanism can be formulated as follows:
\begin{align}
    \mathcal{L}_\mathrm{safe} &= \mathbb{E}_{\tau\sim\pi_\theta}\sum^{T-1}_{t=1}\mathbbm{1}_{\big({f_\mathrm{ttc}}(\tau) < \delta_t\big)}\left\lVert\tau_{t+1} -\mathrm{sg}(\tau_{t})\right\rVert_2^2, \\
    \mathcal{L}_\mathrm{eff} &= \mathbb{E}_{\tau\sim\pi_\theta}\sum^{T-1}_{t=1}\mathbbm{1}_{\big({f_\mathrm{v}}(\tau) < \delta_v\big)}\left\lVert\tau_{t} -\mathrm{sg}(\tau_{t+1})\right\rVert_2^2, \\
    \mathcal{L}_\mathrm{low} &= \mathcal{L}_\mathrm{safe}+\mathcal{L}_\mathrm{eff},
\end{align}
where $\pi_\theta$ denotes the driving policy parameterized by $\theta$, $\tau$ denotes the planned trajectory of length $T$, 
$\tau_t$ denotes the $t$-th trajectory point of $\tau$,
$f_\mathrm{ttc}(\tau)$ and $f_v(\tau)$ represent the TTC and velocity indicators, $\delta_t$ and $\delta_v$ are their respective thresholds, $\mathrm{sg}(\cdot)$ is the stop-gradient operator, and $\mathbbm{1}_{(\cdot)}$ is the indicator function.

\paragraph{High-Level Decision Alignment.}
For the high level, we align the VLM decision with the optimized low-level planning.
We map the refined trajectory to its corresponding high-level decision via the kinematic mapping function $f_K$.
Suppose the VLM decision is inconsistent with the trajectory-corresponding decision. In that case, we further penalize the probability distribution of the VLM decision to ensure dual-system correspondence:
\begin{equation}
    \mathcal{L}_\mathrm{high} = -\log P(f_K(\tau)\mid Q),
\end{equation}
where $d$ denotes the VLM decision and $Q$ denotes the question input.
The final loss function can be formulated as:
\begin{equation}
    \mathcal{L}_\mathrm{stage3} = \mathcal{L}_\mathrm{high} + \beta\mathcal{L}_\mathrm{low},
    \label{eq:stage3}
\end{equation}
where $\beta$ is a balancing weight between high-level and low-level losses.

\begin{table*}[t]
    \centering
    \tablestyle{17.2pt}{1.0} 
    \caption{\textbf{Comparison of decision-planning consistency.} \thename{} exhibits higher consistency between the VLM decision and the E2E planning compared to Senna~\cite{senna}.}
    \begin{tabular}{lcccccccc}
        \toprule
        \multirow{2}{*}[-0.5ex]{Method} & \multicolumn{3}{c}{Path (F1) $\uparrow$} & \multicolumn{4}{c}{Speed (F1) $\uparrow$} & \multirow{2}{*}[-0.5ex]{Avg.}     \\
        \cmidrule(lr){2-4} \cmidrule(lr){5-8}
                           & Straight       & Left           & Right          & Keep           & Acc.           & Dec.           & Stop           &                \\
        \midrule
        Senna~\cite{senna} & 0.763          & 0.533          & 0.574          & 0.550          & 0.612          & 0.628          & 0.802          & 0.637          \\
        \rowcolor{gray!12}
        \textbf{Ours}      & \textbf{0.809} & \textbf{0.664} & \textbf{0.710} & \textbf{0.754} & \textbf{0.769} & \textbf{0.780} & \textbf{0.838} & \textbf{0.760} \\
        \bottomrule
    \end{tabular}\vspace{-1.2em}
    \label{tab:open-loop-consistency}
\end{table*}

\section{Experiments}
\label{sec:experiments}

\subsection{Experimental Settings}

\paragraph{Dataset.}
We collect $\approx 10,000$ hours (360M frames) of expert human driving demonstrations in real-world environments. 
Each demonstration includes synchronized multi-view videos, odometry data with detailed annotations of maps, traffic agents, and navigation information.
Based on the odometry data, we generate high-level decision labels for first- and second-stage training.
In the third stage, we select 1,300 high-risk driving clips, ranging from 15 to 40 seconds in length, and reconstruct them into 3DGS environments. We split 1044 clips for training and 256 clips for closed-loop evaluation.

\paragraph{Metrics.}
We thoroughly evaluate \thename{}'s performance across three aspects: decision-planning consistency, open-loop metrics, and closed-loop metrics. Decision-planning consistency and open-loop metrics are evaluated on a subset of 100-hour driving data. Closed-loop metrics are evaluated in 3DGS environments.

To evaluate decision-planning consistency, we compute the \textbf{F1 score} between the VLM decisions and those obtained from the E2E trajectories via kinematic mapping.

\begin{table}[t]
    \centering
    \tablestyle{1.3pt}{1.0}
    \caption{\textbf{Open-loop quantitative comparisons with existing method.}
    Our method outperforms existing methods in both displacement error and collision rate metrics.}
    \begin{tabular}{lccccc}
        \toprule
        Method & FDE (m) $\downarrow$ & ADE (m) $\downarrow$ & CR (\%) $\downarrow$ & DCR (\%) $\downarrow$ & SCR (\%) $\downarrow$ \\
        \midrule
        TransFuser~\cite{transfuser} & 0.844          & 0.297          & 0.981          & 0.827          & 0.154          \\
        VAD~\cite{vad}               & 0.722          & 0.262          & 0.621          & 0.554          & 0.067          \\
        GenAD~\cite{genad}           & 0.806          & 0.290          & 0.520          & 0.491          & 0.030          \\
        ResAD~\cite{resad}           & 0.634          & 0.234          & 0.378          & 0.367          & 0.011          \\
        Senna~\cite{senna}           & 0.633          & 0.236          & 0.294          & 0.286          & 0.008          \\
        \rowcolor{gray!12}
        \textbf{Ours}                & \textbf{0.597} & \textbf{0.225} & \textbf{0.288} & \textbf{0.283} & \textbf{0.005} \\
        \bottomrule
    \end{tabular}\vspace{-0.7em}
    \label{tab:ol-comparison}
\end{table}

The open-loop evaluation metrics include: 
\textbf{Final Displacement Error (FDE)}: The Euclidean distance between the predicted trajectory endpoint and the ground-truth endpoint;
\textbf{Average Displacement Error (ADE)}: The average distance error between the predicted and ground-truth trajectories over the entire time horizon;
\textbf{Collision Rate (CR)}: The proportion of predicted trajectories that collide with other traffic participants or static obstacles in the scene;
\textbf{Dynamic Collision Rate (DCR)}: The proportion of collisions involving only dynamic objects (\eg, vehicles or pedestrians);
and \textbf{Static Collision Rate (SCR)}: The proportion of collisions involving only static objects (\eg, curbs).

The closed-loop evaluation metrics include:
\textbf{At-fault Collision Rate (AF-CR)}: The proportion of driving clips where collisions occur due to the ego vehicle’s inappropriate decisions;
\textbf{Collision Rate (CR)}: The overall proportion of driving clips where collisions occur during the whole rollout;
and \textbf{Safety@1 / Safety@2}: The proportion of safe driving clips where the minimum value of time-to-collision (TTC) with surrounding agents during the whole rollout exceeds 1 or 2 seconds, respectively.

\begin{table}[t]
    \centering
    \tablestyle{6.5pt}{1.0}
    \caption{\textbf{Closed-loop quantitative comparisons with existing methods on the 3DGS evaluation benchmark.}
    Our method significantly improves safety and reduces the collision rate.}
    \begin{tabular}{lcccccccccccc}
        \toprule
        Method                       & CR $\downarrow$ & AF-CR $\downarrow$ & Safety@1 $\uparrow$ & Safety@2 $\uparrow$ \\
        \midrule
        TransFuser~\cite{transfuser} & 0.435           & 0.269              & 0.531               & 0.454               \\
        VAD~\cite{vad}               & 0.502           & 0.280              & 0.458               & 0.362               \\
        GenAD~\cite{genad}           & 0.557           & 0.244              & 0.402               & 0.332               \\
        ResAD~\cite{resad}           & 0.509           & 0.288              & 0.469               & 0.399               \\
        VADv2~\cite{vadv2}           & 0.422           & 0.199              & 0.514               & 0.458               \\
        RAD~\cite{rad}               & 0.281           & 0.113              &  0.613              & 0.543               \\
        Senna~\cite{senna}           & 0.310           & 0.111              & 0.638               & 0.539               \\
        \rowcolor{gray!20}
        \textbf{Ours}                & \textbf{0.269}  & \textbf{0.077}     & \textbf{0.667}      & \textbf{0.565}      \\
        \bottomrule
    \end{tabular}\vspace{-0.8em}
    \label{tab:cl-comparison}
\end{table}

\subsection{Main Results}

\vspace{-0.1em}\subsubsection{Comparisons with Existing Methods}

We evaluate \thename{} against state-of-the-art methods in both open-loop and closed-loop settings, with all models trained on the same dataset for a fair comparison.
As shown in~\cref{tab:ol-comparison}, \thename{} reduces FDE by 5.7\% over prior methods in the open-loop evaluation.
Closed-loop results in~\cref{tab:cl-comparison} further confirm its effectiveness, yielding a 30.6\% decrease in AF-CR.
Notably, in closed-loop evaluation, \thename{} outperforms RL-based baselines, \eg, RAD~\cite{rad}, despite both employing closed-loop reinforcement training.
This indicates that the gains stem from our alignment strategy rather than closed-loop reinforcement alone.
Overall, these results demonstrate that our method consistently enhances prediction accuracy and driving safety in real-world scenarios.

\vspace{-0.1em}\subsubsection{Decision-Planning Consistency}

We perform both quantitative and qualitative analyses to assess the dual-system consistency between VLM decisions and E2E planning.

\begin{figure*}[t]
    \centering
    \includegraphics[width=0.93\textwidth]{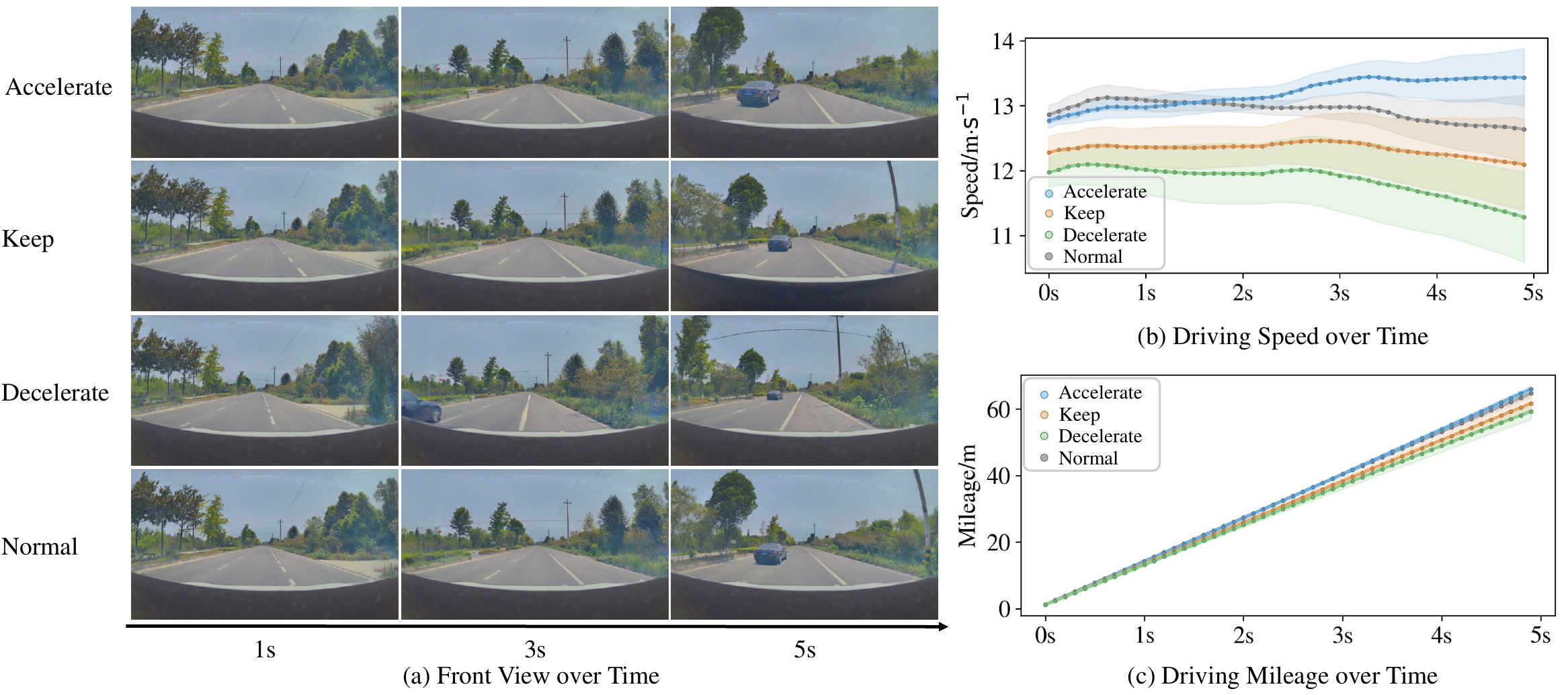}
    \caption{\textbf{Closed-loop speed control in an empty-road scenario.} We visualize (a) driving trajectories, (b) speed curves, and (c) mileage curves under different VLM decisions. Our method exhibits strong decision-following ability. 
    The low-level planning follows the high-level decision for speed control.
    \textit{Normal} denotes using the VLM-predicted decision, while \textit{accelerate}, \textit{keep} and \textit{decelerate} denotes using the fixed ones during the whole rollout.}\vspace{-1.2em}
    \label{fig:closed-loop-consistency}
\end{figure*}

\vspace{-0.8em}\paragraph{Quantitative Results.}
As shown in~\cref{tab:open-loop-consistency}, we compare the decision-planning consistency between our method and Senna~\cite{senna}.
Our approach achieves a 19.3\% improvement in average F1 score, indicating that it significantly improves the alignment between the VLM and the E2E policy.
To intuitively illustrate open-loop controllability, we visualize the relative speed distributions under fixed-speed control actions in~\cref{fig:teaser}.
As shown in~\cref{fig:teaser}(b), Senna exhibits limited separability among the three control modes, indicating weak control differentiation.
In contrast, \thename{} (\cref{fig:teaser}(c)) produces clearly separated distributions, demonstrating more consistent top-down guidance.

\paragraph{Qualitative Analysis.}
We present a closed-loop case study that qualitatively examines the behaviors under different fixed decisions. We perform multiple rollouts with different decisions in the same 3DGS environment. The visualization of the front-view scenes with the corresponding velocity and mileage curves is shown in ~\cref{fig:closed-loop-consistency}.
Our method produces stable and clearly differentiated control profiles for each action, demonstrating effective and interpretable planning consistency in real-world driving scenarios.

\subsection{Ablation Study}

\begin{table}[t]
    \centering
    \tablestyle{2.8pt}{1.0}
    \caption{\textbf{Ablation study on the model architecture.} Dual system: whether the VLM is integrated; VLM token / Dec. token: whether the VLM token/decision token is used in the decision adapter.}
    \begin{tabular}{cccccccc}
        \toprule
        \multirow{2}{*}[0ex]{\makecell[c]{Dual\\system}} & \multirow{2}{*}[0ex]{\makecell[c]{VLM\\token}} & \multirow{2}{*}[0ex]{\makecell[c]{Dec.\\token}} & \multirow{2}{*}[0ex]{\makecell[c]{Open-loop\\FDE (m) $\downarrow$}} & \multirow{2}{*}[0ex]{\makecell[c]{Open-loop\\CR (\%) $\downarrow$}} & \multirow{2}{*}[0ex]{\makecell[c]{Closed-loop\\AF-CR $\downarrow$}} & \multirow{2}{*}[-0.25ex]{\makecell[c]{Avg. F1} $\uparrow$} \\\\
        \midrule
        \ding{55} & \ding{55} & \ding{55} & 0.695          & 0.387          & 0.288          & -              \\
        \ding{51} & \ding{55} & \ding{51} & 0.634          & 0.299          & 0.148          & 0.660          \\
        \ding{51} & \ding{51} & \ding{55} & 0.624          & 0.312          & 0.151          & 0.688          \\
        \rowcolor{gray!12}
        \ding{51} & \ding{51} & \ding{51} & \textbf{0.567} & \textbf{0.281} & \textbf{0.144} & \textbf{0.701} \\
        \bottomrule
    \end{tabular}\vspace{-0.7em}
    \label{tab:arch-ablation}
\end{table}

\begin{table}[t]
    \centering
    \tablestyle{1.8pt}{1.0}
    \caption{\textbf{Ablation study on the training stages.} Our three-stage training pipeline strikes a balance between open-loop and closed-loop performance, while also enhancing the consistency between high-level decisions and low-level planning.}
    \begin{tabular}{ccccccc}
        \toprule
        \multirow{2}{*}[0.25ex]{\makecell[c]{Stage 1}} & \multirow{2}{*}[0.25ex]{\makecell[c]{Stage 2}} & \multirow{2}{*}[0.25ex]{\makecell[c]{Stage 3}} & \multirow{2}{*}[0ex]{\makecell[c]{Open-loop\\FDE (m) $\downarrow$}} & \multirow{2}{*}[0ex]{\makecell[c]{Open-loop\\CR (\%) $\downarrow$}} & \multirow{2}{*}[0ex]{\makecell[c]{Closed-loop\\AF-CR $\downarrow$}} & \multirow{2}{*}[-0.25ex]{\makecell[c]{Avg. F1} $\uparrow$} \\\\
        \midrule
        \ding{51} &                               &           & \textbf{0.567} & \textbf{0.281}  & 0.144           & 0.701 \\
        \ding{51} & \ding{51}                     &           & 0.575                       & 0.290                        & 0.118           & \textbf{0.764}     \\
        \rowcolor{gray!12}
        \ding{51} & \ding{51}                     & \ding{51} & 0.597                       & 0.288                        & \textbf{0.077}  & 0.760              \\
        \bottomrule
    \end{tabular}\vspace{-0.7em}
    \label{tab:training}
\end{table}

\begin{table*}[t]
    \centering
    \tablestyle{10.5pt}{1.0}
    \caption{\textbf{Performance on the NAVSIM v2 \texttt{navtest} Benchmark.}}
    \begin{tabular}{lcccccccccc}
        \toprule
        Method                               & NC $\uparrow$ & DAC $\uparrow$ & DDC $\uparrow$ & TL $\uparrow$ & EP $\uparrow$ & TTC $\uparrow$ & LK $\uparrow$ & HC $\uparrow$ & EC $\uparrow$ & EPDMS $\uparrow$ \\
        \midrule
        Transfuser~\cite{transfuser}         & 96.9          & 89.9           & 97.8           & 99.7          & 87.1          & 95.4           & 92.7          & 98.3          & 87.2          & 76.7             \\
        Hydra-MDP++~\cite{hydramdp++}        & 97.2          & 97.5           & 99.4           & 99.6          & 83.1          & 96.5           & 94.4          & 98.2          & 70.9          & 81.4             \\
        DriveSuprim~\cite{drivesuprim}       & 97.5          & 96.5           & 99.4           & 99.6          & \textbf{88.4} & 96.6           & 95.5          & 98.3          & 77.0          & 83.1             \\
        ARTEMIS~\cite{artemis}               & 98.3          & 95.1           & 98.6           & \textbf{99.8} & 81.5          & 97.4           & 96.5          & 98.3          & -             & 83.1             \\
        DiffusionDrive~\cite{diffusiondrive} & 98.2          & 95.9           & 99.4           & \textbf{99.8} & 87.5          & 97.3           & 96.8          & 98.3          & 87.7          & 84.5             \\
        ResAD~\cite{resad}                   & 97.8          & 97.2           & \textbf{99.5}  & \textbf{99.8} & 88.2          & 96.9           & \textbf{97.0} & 98.4 & 88.2 & 85.5             \\
        \rowcolor{gray!12}
        \textbf{Ours}                        & \textbf{98.5} & \textbf{97.8}  & \textbf{99.5}  & \textbf{99.8} & 88.1          & \textbf{97.5}  & \textbf{97.0} & \textbf{98.6} & \textbf{88.4}          & \textbf{86.6}    \\
        \bottomrule
    \end{tabular}\vspace{-0.7em}
    \label{tab:navsim}
\end{table*}

\begin{figure*}[t]
    \centering
    \includegraphics[width=0.96\textwidth]{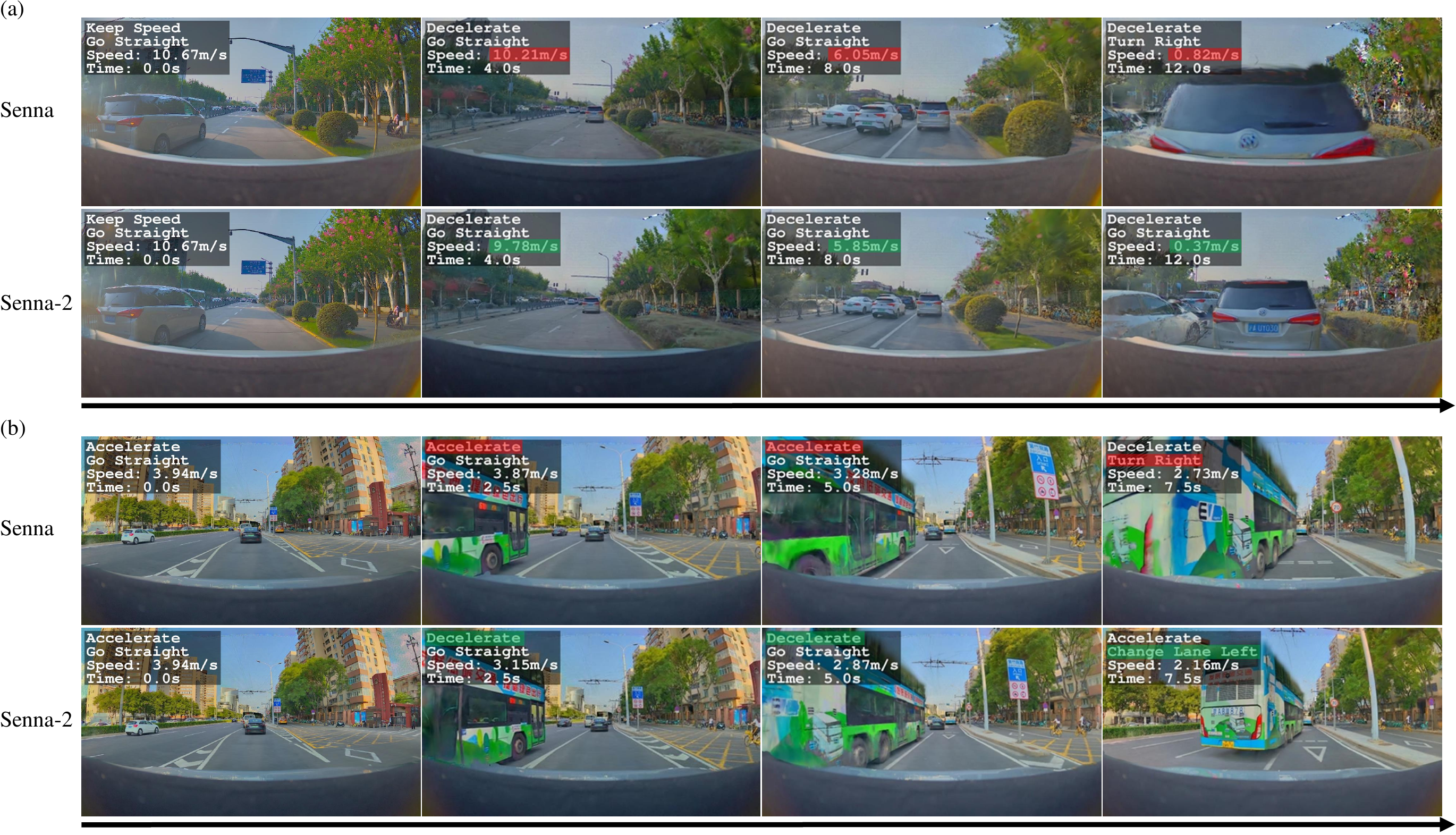}
    \caption{\textbf{Closed-loop qualitative comparisons between Senna~\cite{senna} and \thename{}.}
Senna suffers from (a) planning misalignment in \textit{stop} scenario and (b) decision misalignment in \textit{cut-in} scenario.
In contrast, \thename{} maintains both consistent decision-making and accurate trajectory planning across these challenging situations, demonstrating improved safety and reliability in closed-loop driving.}
    \label{fig:qualitative}\vspace{-0.7em}
\end{figure*}

\paragraph{Model Architecture.}
As illustrated in~\cref{tab:arch-ablation}, we present an ablation study on the model architecture to further validate the contributions of the VLM and the decision adapter.
All experiments in this table are trained using only Stage 1 to ensure a fair and independent comparison.

To evaluate the effectiveness of the dual-system design, we compare the pure E2E baseline without VLM guidance (row 1) and the proposed dual-system model (row 4).
\thename{} consistently improves safety and planning accuracy, yielding a 27.4\% reduction in collision rate and a 0.128-meter decrease in FDE.
This confirms that high-level decisions provided by the VLM offer reliable guidance that enhances the stability and safety of the E2E planner.

We further ablate the decision adapter by removing the VLM token (row 2) and decision tokens (row 3).
Excluding the VLM token leads to clear degradation (11.8\% FDE increase and 6.4\% CR increase), while removing the decision token also harms performance (10.1\% FDE increase and 11.0\% CR increase).
These results show that VLM semantics and explicit awareness of decision categories are both essential for accurate and consistent planning.

\paragraph{Training Stage.}
We further present an ablation study validating the effectiveness of our proposed training recipe in~\cref{tab:training}.
Training with only Stage 1 yields the lowest open-loop FDE, as the model is optimized purely for trajectory regression without considering dual-system interactions.
However, this also leads to lower consistency and weaker closed-loop performance.
Introducing Stage 2 significantly strengthens dual-system consistency with a 9.0\% F1 score improvement, demonstrating the significance of explicit alignment.
Incorporating the full Stage 3 directly improves closed-loop behavior with a 34.7\% AF-CR reduction, leading to better generalization in open-world driving scenarios.
These results show that the three stages play complementary roles and the complete pipeline achieves a balance between open-loop accuracy, closed-loop robustness, and decision–planning consistency.

\subsection{Performance on Open-Source NAVSIM v2 Benchmark}
We further finetune our model on the NAVSIM training split~\cite{navsim} and evaluate the planning performance of \thename{} on the NAVSIM v2 benchmark~\cite{navsimv2}.
As demonstrated in \cref{tab:navsim}, our approach yields an EPDMS of 86.6, which surpasses the previous best-performing model by 1.1 EPDMS.
This result underscores the robustness of \thename{} and its capacity to master complex driving scenes.

\subsection{Qualitative Results}
We further present closed-loop qualitative comparisons between \thename{} and Senna~\cite{senna} in~\cref{fig:qualitative}.
In~\cref{fig:qualitative}(a), Senna exhibits a clear planning misalignment: although the VLM issues a correct \textit{decelerate} command, the planner fails to respond in time, ultimately resulting in a rear-end collision with the lead vehicle.
In contrast, \thename{} maintains consistent decision–planning alignment and achieves a safe stop.
In~\cref{fig:qualitative}(b), Senna suffers from decision errors in a cut-in scenario.
It first outputs an incorrect \textit{accelerate} command, causing the planner to miss the proper braking window and collide with the merging vehicle. By 7.5 s, it further issues an unreasonable turn-right action after the collision.
\thename{}, however, consistently provides the correct \textit{deceleration} decision to manage the merge safely, and produces a reasonable \textit{change-line-left} command that completes the avoidance maneuver.
These results indicate that \thename{} ensures more reliable dual-system consistency and safer closed-loop behavior in challenging situations.

\section{Conclusion and Limitation}
\label{sec:conclusion}
In this work, we present \thename{}, a dual-system driving policy that aligns a Vision-Language Model (VLM) with an end-to-end (E2E) driving policy for consistent decision-making and planning.
By introducing a three-stage training strategy comprising Driving Pre-Training, Open-Loop Alignment, and Closed-Loop Alignment with HRL, \thename{} effectively bridges the VLM decision-making with the planning of E2E models.
Extensive experiments on both open-loop and closed-loop benchmarks demonstrate that \thename{} significantly improves driving safety and interpretability over state-of-the-art baselines.
This work highlights the potential of integrating multimodal reasoning into end-to-end driving systems, paving the way toward more reliable, explainable, and human-aligned autonomous driving.

A limitation is that the VLM currently cannot achieve real-time inference (10 Hz) on on-board edge devices.
The VLM and E2E policy operate asynchronously, with a memory bank caching VLM features for dual-system interaction.
Achieving fully synchronous cooperation requires further hardware optimization.

{
    \small
    \bibliographystyle{ieeenat_fullname}
    \bibliography{main}
}

\clearpage
\section*{Appendices}
\setcounter{page}{1}
\setcounter{table}{0}
\renewcommand{\thetable}{A\arabic{table}}%
\setcounter{figure}{0}
\renewcommand{\thefigure}{A\arabic{figure}}%
\setcounter{section}{0}
\renewcommand{\thesection}{A.\arabic{section}}
\section{Training Configurations}
\label{sec:impl_details}
In this section, we summarize the training configurations and hyperparameters used across the three stages of our framework: Driving Pre-training, Open-Loop Alignment, and Closed-Loop Alignment with HRL.

\subsection{Driving Pre-training}
Driving pre-training includes the pre-training of the vision-language model (VLM), the end-to-end model (E2E), and the driving adapter.

\paragraph{VLM Pre-Training.}
We use Qwen2.5VL-3B~\cite{qwen2_5vl} as our base model.
For each driving demonstration, we sample frames at 1 fps.
For every sampled frame, we obtain the corresponding decision label by applying the kinematic mapping function to the expert trajectories.
The detailed training configurations are listed in \cref{tab:vlm_pretrain_cfg}.

\begin{table}[ht]
    \centering
    \tablestyle{9pt}{1.0}
    \caption{Training configurations for VLM pre-training.}
    \label{tab:vlm_pretrain_cfg}
    \begin{tabular}{lc}
    \toprule
    Config                     & Value                                                   \\
    \midrule
    learning rate              & $5\times 10^{-5}$                                       \\
    learning rate schedule     & cosine decay                                            \\
    optimizer                  & AdamW~\cite{adam,adamw}                                 \\
    optimizer hyper-parameters & $\beta_1, \beta_2, \epsilon=0.9, 0.999, 1\times10^{-8}$ \\
    weight decay               & $1\times10^{-4}$                                        \\
    batch size                 & 512                                                     \\
    training steps             & 10k                                                     \\
    training GPU               & 128 NVIDIA L20                                          \\
    \bottomrule
    \end{tabular}
\end{table}

\paragraph{E2E Pre-Training.}
We provide the detailed hyperparameters for E2E pre-training in \cref{tab:e2e_pretrain_cfg}.

\begin{table}[ht]
    \centering
    \tablestyle{9pt}{1.0}
    \caption{Training configurations for E2E pre-training.}
    \label{tab:e2e_pretrain_cfg}
    \begin{tabular}{lc}
    \toprule
    Config                     & Value                                                   \\
    \midrule
    learning rate              & $1\times 10^{-4}$                                       \\
    learning rate schedule     & cosine decay                                            \\
    optimizer                  & AdamW~\cite{adam,adamw}                                 \\
    optimizer hyperparameters & $\beta_1, \beta_2, \epsilon=0.9, 0.999, 1\times10^{-8}$ \\
    weight decay               & $1\times10^{-4}$                                        \\
    batch size                 & 12288                                                     \\
    training steps             & 30k                                                     \\
    training GPU               & 1024 NVIDIA H20                                          \\
    \bottomrule
    \end{tabular}
\end{table}

\paragraph{Adapter Pre-Training.}
We train all parameters of the driving adapter from scratch and finetune all parameters of the E2E planner.
The detailed hyperparameters for adapter pre-training are provided in \cref{tab:adapter_pretrain_cfg}.

\begin{table}[ht]
    \centering
    \tablestyle{9pt}{1.0}
    \caption{Training configurations for adapter pre-training.}
    \label{tab:adapter_pretrain_cfg}
    \begin{tabular}{lc}
    \toprule
    Config                     & Value                                                   \\
    \midrule
    learning rate              & $5\times 10^{-5}$                                       \\
    learning rate schedule     & cosine decay                                            \\
    optimizer                  & AdamW~\cite{adam,adamw}                                 \\
    optimizer hyper-parameters & $\beta_1, \beta_2, \epsilon=0.9, 0.999, 1\times10^{-8}$ \\
    weight decay               & $1\times10^{-4}$                                        \\
    batch size                 & 512                                                     \\
    training steps             & 10k                                                     \\
    training GPU               & 128 NVIDIA L20                                          \\
    \bottomrule
    \end{tabular}
\end{table}

\subsection{Open-Loop Alignment}

In each iteration, we first perform a two-step DDIM rollout to generate the predicted trajectory and compare it with the VLM decision.
Only the inconsistent samples are used for training.
The loss weight $\gamma$ is set to 1.
During optimization, we apply LoRA finetuning~\cite{lora} to the VLM while fully finetuning the remaining modules.
The detailed hyperparameters are listed in \cref{tab:openloop_cfg}.

\begin{table}[ht]
    \centering
    \tablestyle{9pt}{1.0}
    \caption{Training configurations for open-loop alignment.}
    \label{tab:openloop_cfg}
    \begin{tabular}{lc}
    \toprule
    Config                     & Value                                                   \\
    \midrule
    learning rate              & $1\times 10^{-5}$                                       \\
    learning rate schedule     & cosine decay                                            \\
    optimizer                  & AdamW~\cite{adam,adamw}                                 \\
    optimizer hyper-parameters & $\beta_1, \beta_2, \epsilon=0.9, 0.999, 1\times10^{-8}$ \\
    weight decay               & 0                                                       \\
    batch size                 & 384                                                     \\
    training steps             & 5k                                                      \\
    LoRA hyperparameters       & $r, \alpha=8, 16$                                       \\
    training GPU               & 128 NVIDIA L20                                          \\
    \bottomrule
    \end{tabular}
\end{table}

\subsection{Closed-Loop Alignment with HRL}
We assign one 3DGS environment to each GPU and perform rollouts at 10 Hz in every loop.
After each rollout, we compute per-frame TTC and speed.
Frames with a TTC below 3 seconds are regarded as safety-critical,
while frames whose speed and distance fall behind the expert’s are treated as inefficient. 
To avoid introducing noisy data, all frames after the first collision are discarded.
The remaining valid rollout frames are subsampled by taking one frame every four for training.
The loss weight $\beta$ is set to 1.
For model updates, we follow the same strategy as in Stage 2,
applying LoRA finetuning~\cite{lora} to the VLM and full-parameter finetuning to the remaining components.
The detailed hyperparameters are listed in \cref{tab:closedloop_cfg}.

\begin{table}[t]
    \centering
    \tablestyle{9pt}{1.0}
    \caption{Training configurations for closed-loop alignment.}
    \label{tab:closedloop_cfg}
    \begin{tabular}{lc}
    \toprule
    Config                     & Value                                                   \\
    \midrule
    learning rate              & $1\times 10^{-5}$                                       \\
    learning rate schedule     & cosine decay                                            \\
    optimizer                  & AdamW~\cite{adam,adamw}                                 \\
    optimizer hyper-parameters & $\beta_1, \beta_2, \epsilon=0.9, 0.999, 1\times10^{-8}$ \\
    weight decay               & 0                                                       \\
    batch size                 & 128                                                     \\
    training steps             & 500                                                     \\
    LoRA hyperparameters       & $r, \alpha=8, 16$                                       \\
    training GPU               & 64 NVIDIA L20                                           \\
    \bottomrule
    \end{tabular}
\end{table}

\section{Kinematic Mapping Function}
\label{sec:kmf}
In this section, we introduce the implementation details on the kinematic mapping function $f_k(\tau)$.
This function is used to convert trajectories into high-level decision labels for two purposes:
generating ground-truth decisions from expert data, and inferring decisions from E2E planning for consistency checking with VLM outputs.
We describe the function in two parts: speed and direction control.

For speed control, given a trajectory, we first extract the first 1.5s subsequence $\tau=\left\{\tau_t\right\}^{14}_{t=0}$ with a timestep of $\Delta t=0.1$s. The speed $v_t$ and acceleration $a_t$ are computed as
\begin{equation}
    v_t = \frac{\lVert \tau_t - \tau_{t-1} \rVert}{\Delta t}, \quad a_t = \frac{v_t - v_{t-1}}{\Delta t},
\end{equation}
and smooth both with an average filter of window size $w=5$.
Let $f_{\text{lcs}}$ return the length of the longest continuous True segment in a Boolean sequence, and $\text{RMS}(\cdot)$ return the mean square sum.
Then the trajectory is labeled as \textit{accelerate} if
\begin{equation}
    \begin{cases}
        a_1>0,\quad f_{\text{lcs}}(a_t>0.3)\ge8,\\\max(a_t)>0.6,\quad \text{RMS}(a_t)>0.4. 
    \end{cases}
\end{equation}
The trajectory is labeled as \textit{decelerate} if
\begin{equation}
    \begin{cases}
        a_1<0,\quad f_{\text{lcs}}(a_t < -0.3)\ge8,\\\min(a_t)<-0.6,\quad\text{RMS}(a_t)>0.4. 
    \end{cases}
\end{equation}
If neither condition is met, we compute the average speed $\bar v$ and acceleration $\bar a$.
The label is assigned as \textit{keep speed} if
\begin{equation}
    \begin{cases}
        |\bar a|<0.3\cdot s_a,\\\max(|a_t|)<0.6\cdot s_a,
    \end{cases}
\end{equation}
or \textit{stop} if $\bar v<0.5$, where
\begin{equation}
    s_a=\begin{cases}
        2.5,&\bar v>25,\\2,&20<\bar v\le25,\\1.5,&10<\bar v\le20,\\1.25,&5<\bar v\le10,\\1,&otherwise.\\
    \end{cases}
\end{equation}
If none of the above conditions hold, the speed label is considered unknown and ignored.

For direction control, we consider two cases.
For the predicted trajectory, we first compute the average speed $\bar v$, the maximum and minimum lateral displacement $\Delta y_{\text{max}}$ and $\Delta y_{\text{min}}$, and the maximum yaw variation $\Delta \psi_{\text{max}}$ within the 1.5-second horizon, relative to the initial time.
The direction decision label $d^{\text{kmf}}_{\text{dir}}$ is assigned as
\begin{equation}
    d^{\text{kmf}}_{\text{dir}}=\begin{cases}
        \text{\textit{left}},&\text{if } \Delta \psi_{\text{max}}>s_\psi \text{ and } \Delta y_{\text{max}}>s_y,\\
        \text{\textit{right}},&\text{if } \Delta \psi_{\text{max}}>s_\psi \text{ and } \Delta y_{\text{min}}<-s_y,\\
        \text{\textit{staight}},&otherwise,\\
    \end{cases}
\end{equation}
where $s_\psi=\frac{\pi}{36}$, and
\begin{equation}
    s_y=\begin{cases}
        3,&\bar v>15,\\2.4,&10<\bar v\le15,\\1.5,&5<\bar v\le10,\\0.9,&3<\bar v\le5,\\0.45,&otherwise.\\
    \end{cases}
\end{equation}
For VLM pre-training, we employ frame-level expert annotations $d^{\text{gt}}_{\text{dir}}$ as ground truth. These annotations encompass multiple categories, including \textit{go straight}, \textit{turn left}, \textit{turn right}, \textit{change lane left}, and \textit{change lane right}.
During consistency checking, the fine-grained VLM decisions are projected onto the coarse decision classes used by the E2E policy as follows:
\begin{equation}
    d^{\text{vlm}}_{\text{dir}}=\begin{cases}
        \text{\textit{left}},&\text{if } d^{\text{gt}}_{\text{dir}}\in\left\{\text{\textit{turn left}, \textit{change lane left}}\right\},\\
        \text{\textit{right}},&\text{if } d^{\text{gt}}_{\text{dir}}\in\left\{\text{\textit{turn right}, \textit{change lane right}}\right\},\\
        \text{\textit{staight}},&otherwise.\\
    \end{cases}
\end{equation}

\begin{figure}[t]
    \centering
    \includegraphics[width=\columnwidth]{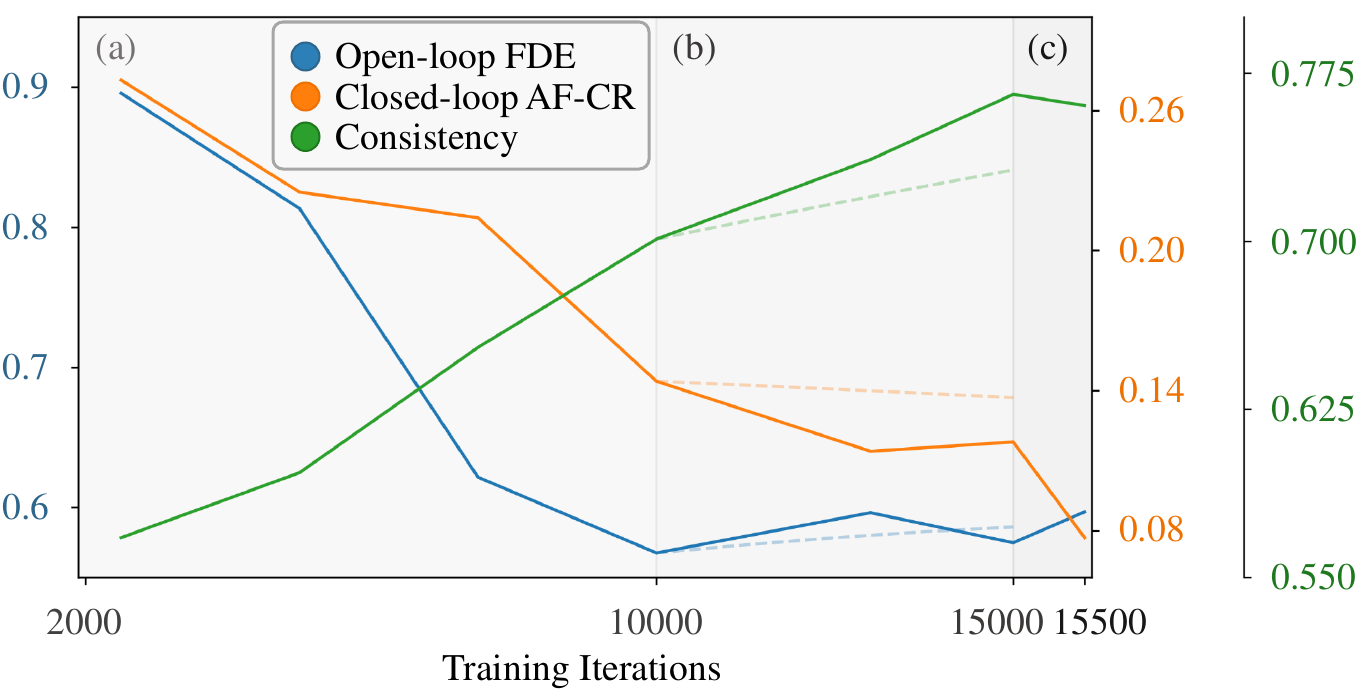}
    \caption{\textbf{Learning curve across the whole training pipeline.} (a) Driving pretraining. (b) Open-loop alignment. (c) Closed-loop alignment with HRL. The dashed line denotes the performance of the extended driving-pretraining baseline.}
    \label{fig:dynamics}
\end{figure}

\begin{figure*}[t]
    \centering
    \includegraphics[width=\textwidth]{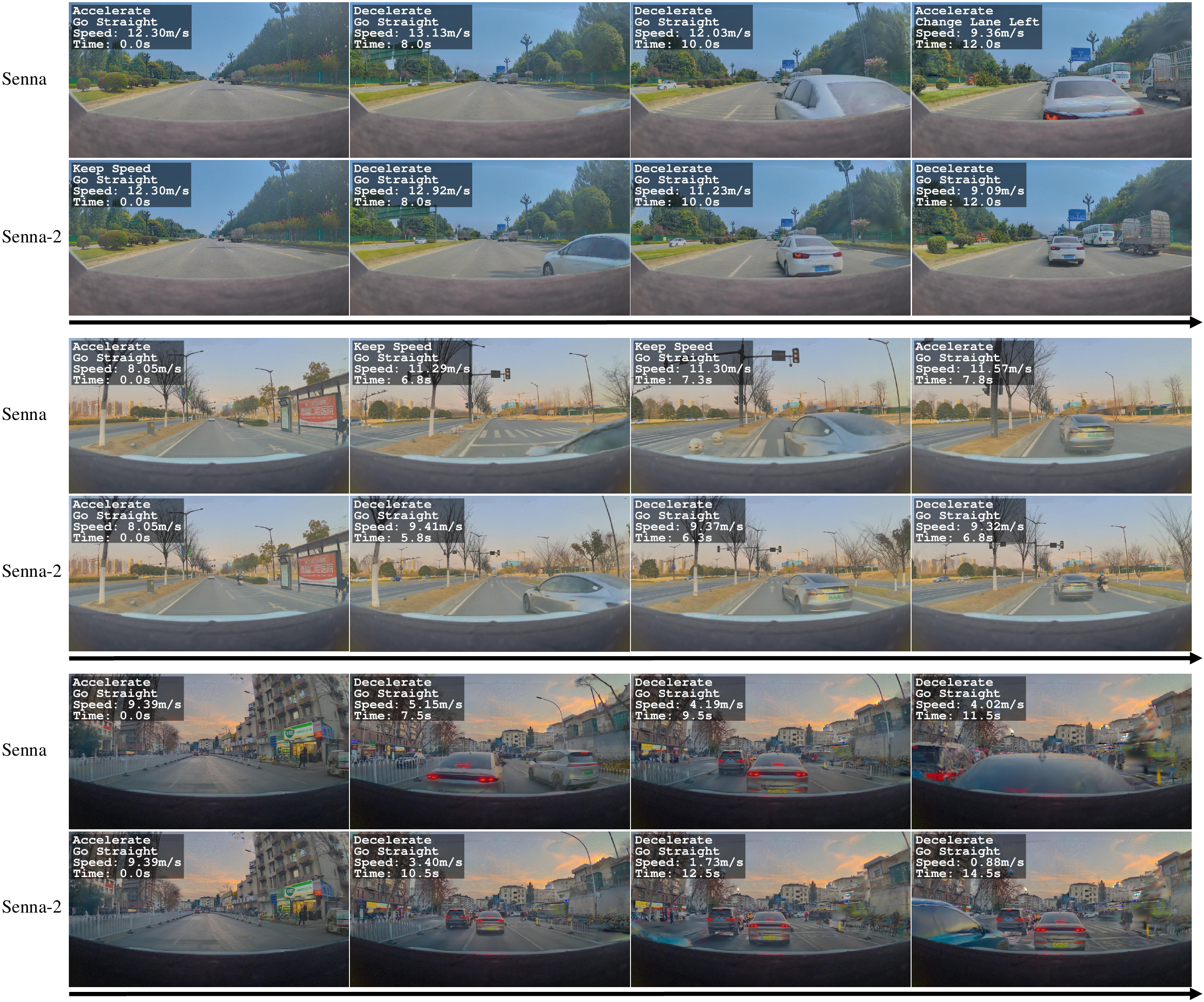}
    \caption{\textbf{More qualitative results in the cut-in scenarios.} In these cases, Senna~\cite{senna} often reacts too late and leads to near-collision outcomes, while our policy maintains coherent decision–planning coupling and ensures safe avoidance.}
    \label{fig:more_vis1}
\end{figure*}

\section{Training Dynamics}
\label{sec:training_dynamics}
As illustrated in \cref{fig:dynamics}, the learning curves highlight the effectiveness of our three-stage pipeline.
Driving pretraining enables the model to acquire core driving competencies from large-scale demonstrations efficiently.
Open-loop alignment substantially enhances decision–planning consistency, delivering greater performance gains than simply prolonging pretraining.
Finally, closed-loop alignment with HRL requires only limited additional training yet produces marked improvements in interactive driving performance.

\section{More Qualitative Results}
\label{sec:more_qual}
\begin{figure*}[t]
    \centering
    \includegraphics[width=\textwidth]{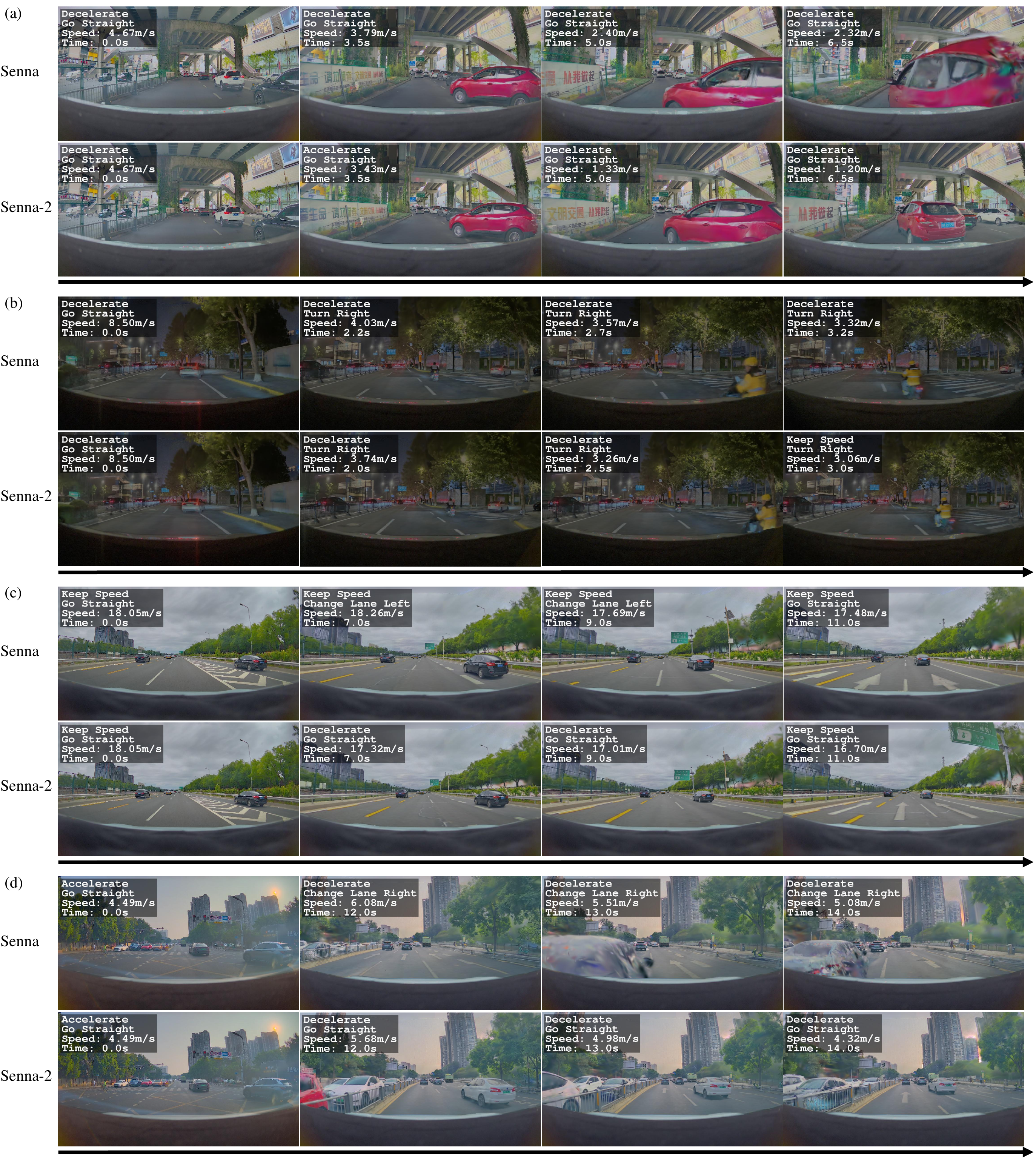}
    \caption{\textbf{More qualitative results in diverse scenarios.}
(a) Our policy safely handles a hard-braking event, while the baseline causes a collision.
(b) During a right turn, our method avoids the cyclist, whereas the baseline crashes.
(c) Our policy completes the lane change, but the baseline drifts across lanes.
(d) Our method aborts the lane change upon hazard detection, while the baseline is rear-ended.}
    \label{fig:more_vis2}
\end{figure*}

We present additional visualizations in \cref{fig:more_vis1} and \cref{fig:more_vis2} that compare our method against the baseline~\cite{senna} across a wide range of scenarios, including cut-in, right-turn, lane-change, and hard-braking events.
These qualitative results highlight the performance gains and improved consistency achieved by our approach under diverse driving conditions, particularly in high-risk interactions where the baseline often reacts late or produces inconsistent planning, leading to near-collision behaviors.
Corresponding video visualizations are included in the supplementary material for clearer dynamic interpretation.

\end{document}